\documentclass[letterpaper]{article}

\usepackage[pass]{geometry}
\usepackage{graphicx}

\usepackage[cmex10]{amsmath}
\usepackage{amssymb}

\usepackage{algorithmic}

\usepackage{url}

\usepackage{epstopdf} 
\begin{document}
%
\title{Unsupervised Video Analysis Based on a Spatiotemporal Saliency Detector}

\author{Qiang~Zhang,
        Yilin~Wang,
        and~Baoxin~Li, \\
 Department of Computer Science, Arizona State University, Tempe AZ \\
 qzhang53,ywang370,baoxin.li@asu.edu}
\maketitle

\begin{abstract}
Visual saliency, which predicts regions in the field of view that draw the most visual attention, has attracted a lot of interest from researchers. It has already been used in several vision tasks, e.g., image classification, object detection, foreground segmentation. Recently, the spectrum analysis based visual saliency approach has attracted a lot of interest due to its simplicity and good performance, where the phase information of the image is used to construct the saliency map. In this paper, we propose a new approach for detecting spatiotemporal visual saliency based on the phase spectrum of the videos, which is easy to implement and computationally efficient. With the proposed algorithm, we also study how the spatiotemporal saliency can be used in two important vision task, abnormality detection and spatiotemporal interest point detection. The proposed algorithm is evaluated on several commonly used datasets with comparison to the state-of-art methods from the literature. The experiments demonstrate the effectiveness of the proposed approach to spatiotemporal visual saliency detection and its application to the above vision tasks.
\end{abstract}

\section{Introduction}
In the recent years modeling and detection of visual saliency has attracted a lot of interest in the vision community. One early work that is widely known is the approach by Itti \emph{et al.} \cite{itti1998model}. Since then, a lot of different models have been proposed for computing visual saliency. Such models may be roughly divided into two groups: bottom-up models (or stimulus driven) that are mainly based on low-level visual features of the scene, and top-down model (goal-driven) that employs information and knowledge about a visual task. A survey of both groups of methods was reported in \cite{borji2012state}. Visual saliency analysis has been applied with success to other vision tasks including object detection \cite{Alexe2012objectness}, image classification \cite{sharma2012discriminative} and foreground segmentation \cite{li2011saliency}.

Recently, spectral-based approach has gained increased interest due to its simplicity and good performance. In \cite{hou2007saliency}, the spectrum residual together with the phase information was used to construct a saliency map. In \cite{guo2008spatio} it was found that it is the phase information rather than the spectrum leads to a better saliency map. 
However, there was a lack of theoretic justification for such methods until \cite{hou2012image}, where it was shown that, if the background is sparsely supported in the DCT domain and the foreground is sparsely supported in the spatial domain
the foreground will receive high value on the computed saliency map.

In the real world, the visual field-of-view of a human may constantly change, and thus visual saliency often depend on not only a static scene but also the changes in the scene. To this end, spatiotemporal saliency has been proposed, which tries to capture regions attracting visual attention in the spatiotemporal domain. Spatiotemporal saliency has been applied to vision tasks such as video summarization \cite{ma2005generic}, human-computer interaction \cite{itti2004realistic}, video compression \cite{guo2010novel}, and abnormality detection \cite{gao2009discriminant}.

In this paper we propose a novel spatiotemporal visual saliency detector for video analysis, based on \emph{the phase information of the video}. With the saliency map computed using the proposed method, we analysis how it can be used for two fundamental vision tasks, namely abnormality detection and spatiotemporal interest point detection. We evaluate the performance of the proposed algorithm using several widely used datasets, with the comparison to the state-of-art in the literature.

The proposed method, compared with the existing work on spatiotemporal saliency in the literature, has several advantages. First, it computes the saliency information from the entire video span, which is different from many existing approaches in the literature. For example, \cite{guo2008spatio} computes temporal information by only the differences of two adjacent frames, which is insufficient for modeling complex activities, as shown in the experiment part. Second, the proposed approach is easy to implement and computationally efficient. The core parts of the algorithm involve only a three-dimensional Fourier transform, whose complexity is only $O(N \log{N})$, where $N$ is the size of the input. Last but not least, no training stage or prior information is needed for the proposed approach, which is a significant advantage for applications like abnormality detection.

The rest of the paper is organized as follows: in Sec. \ref{sec:proposed} we describe the proposed method including the analysis and the relationships with the existing methods; Sec. \ref{sec:experiment} evaluate the proposed spatiotemporal saliency detector in saliency detection on both synthetic dataset and two real video dataset; studies of how the spatiotemporal saliency computed by the proposed method can be used for two important vision tasks, abnormality detection and spatiotemporal interst point detection, is presented in Sec. \ref{sec:application}; and the paper is concluded in Sec. \ref{sec:conclusion}.
\section{Proposed Method}\label{sec:proposed}
As reviewed above, spectrum analysis based approaches to visual saliency has seen some success, although the existing work has been primarily on predicting salient objects for a given (static) image. For a dynamic scene, temporal information should be taken into consideration for properly predicting the salient objects. For example, it was found in \cite{olveczky2003segregation} that objects attract more visual attention if they move differently than their neighbors. Considering this, we propose to compute the saliency map of dynamic scenes by utilizing the phase information of the temporal domain together with the phase information of the spatial domain. In the proposed method, we compute the saliency map for 3D data $\textbf{X}\in\mathbb{R}^{M\times N\times T}$ as:
\begin{equation}\label{eqn:base}
\textbf{Z}=\left\lvert\mathcal{F}^{-1}\left(\frac{\textbf{Y}}{|\textbf{Y}|}\right)\right\rvert^2
\end{equation}
where $\textbf{Y}=\mathcal{F}(\textbf{X})$, $\mathcal{F}$ is 3D discrete Fourier transform and $\mathcal{F}^{-1}$ is the corresponding inverse transform. After we get the saliency map, we smooth it with a 3D Gaussian smooth filter. The 3D Fourier transform can be computed as:
\begin{eqnarray}
&&\textbf{Y}(u,v,w)\\\nonumber
&=&\sum_t{\sum_i{\sum_j{\textbf{X}(i,j,t)e^{-i2\pi\left(\frac{ui}{M}+\frac{vj}{N}+\frac{wt}{T}\right)}}}}\\\nonumber
&=&\sum_t{e^{-i2\pi\frac{wt}{T}}\sum_i{\sum_j{\textbf{X}(i,j,t)e^{-i2\pi\left(\frac{ui}{M}+\frac{vj}{N}\right)}}}}\\\nonumber
&=&\sum_t{e^{-i2\pi\frac{wt}{T}}\sum_i{e^{-i2\pi\frac{ui}{M}}\sum_j{\textbf{X}(i,j,t)e^{-i2\pi\frac{ui}{M}}}}}
\end{eqnarray}
i.e., the 3D Fourier transform can be computed as a sequence of 1D Fourier transforms on each coordinate.

The proposed method detects spatiotemporal saliency, which has been also discussed in some existing works. For example, in \cite{guo2008spatio}, the detection was done by combining color information of one frame and the differences of this frame to the previous one with quaternion Fourier transform. As a result, the temporal information is limited to two adjacent frames and is insufficient for modeling complex scenes. On the other hand, the spatiotemporal saliency proposed in this paper considers the temporal information over a long temporal span up to the entire video.

The method in Eqn. \ref{eqn:base} evaluates the saliency of a region by exploring the information of the entire video. In some situations, we may also be interested in detecting a region that is salient within a temporal window of the video. For example, if a video contains multiple scenes, each capturing a different activity, we may be more interested in analysis the saliency within each scene instead of the entire video. For this reason, we propose multiscale analysis for spatiotemporal saliency, which is inspired by short-time Fourier transform. We first apply the window function to the input signal $\textbf{X}\cdot\textbf{w}(i,j,t)$, where $\cdot$ is the element-wise multiplication and $\textbf{w}(i,j,t)$ the window function centered at position $(i,j,t)$, which is nonzero for only a small support (i.e., the size of window function). The saliency map is computed for the windowed signal:
\begin{eqnarray}
\textbf{Y}&=&\mathcal{F}[\textbf{X}\cdot\textbf{w}(i,j,t)]\\\nonumber
\textbf{Z}(i,j,t)&=&\mathcal{F}^{-1}\left[\frac{\textbf{Y}}{|\textbf{Y}|}\right]
\end{eqnarray}
By sliding the window function on the input video, we still obtain the saliency map for the entire video. The size of the sliding window determines the temporal resolution: with a larger window, more global information of the input is revealed but the resolution is lower; with a smaller window, resolution is improved. The window function can be applied in either temporal direction, spatial direction or both. As a result, we can perform saliency detection from varying scales, which enables us to reveal the information at different spatiotemporal resolution, similar to short time Fourier transform.

Combining different visual cues is important for not only scene saliency but also spatiotemporal saliency. In this paper, we proposed to compute the saliency map for each cue independently then compute the summation of saliency maps from all visual cues. In \cite{guo2010novel}, quaternion Fourier transform (QFT) is utilized to combine the three-channel color information and frame differences. However, the QFT could be very expensive (e.g., time consuming) when applied in spatiotemporal domain. In fact we find that (Appendix \ref{sec:qft}): given a data with four feature channels, the saliency map computed with QFT is very similar with the sum of saliency maps computed with FFT over each feature channel.

Finally, we summarize the proposed algorithm below:\\
\rule{8.4cm}{.2pt}\\
\textbf{Algorithm}\\
\textbf{Input}: data $\textbf{X}$, Gaussian filter $g$, window function $\textbf{w}$\\
\textbf{Output}: saliency map $\textbf{Z}$\\
\textbf{For} each window location\\
\indent\textbf{For} each feature channel\\
\indent\indent Apply $\textbf{w}$ to the input $\textbf{X}$;\\
\indent\indent Compute Fourier coefficient $\textbf{Y}=\mathcal{F}[\textbf{X}]$;\\
\indent\indent Extract the phase information $\hat{\textbf{Y}}=\frac{\textbf{Y}}{|\textbf{Y}|}$;\\
\indent\indent Do the inverse transform $\textbf{Z}=\left\lvert \mathcal{F}^{-1}[\hat{\textbf{Y}}]\right\rvert^2$;\\
\indent\indent Smooth saliency map $\textbf{Z}=\textbf{Z}*g$;\\
\indent\textbf{End}\\
\indent Combine the $\textbf{Z}$ of all channels together;\\
\textbf{End}\\
\rule{8.4cm}{.2pt}\\
where $\textbf{W}$ is the window function. Currently, we only apply the window function along temporal direction and rectangle window is used. The size of the window is depending on the data. By incorporating the phase information of the temporal domain, the proposed method can not only suppress the background, as achieved by visual saliency for images, but also suppress the object which is static or moving ``regularly".
\subsection{Analysis}
There has been several explanations for why spectral domain based approach is able to detect saliency region from the image. For example, \cite{bian2009biological} explained by its biological plausibility that saliency map exists in the primary visual cortex (V1), which is orientation selective and lateral surround inhibition \cite{simoncelli99modelingsurround}. The spectral magnitude measures the total response of cells tuned to the specific frequency and orientation. According to lateral surround inhibition, similarly tuned cells will be suppressed depending on their total response, which can be modeled by dividing its spectral by the spectral magnitude \cite{zhaoping2006pre}. \cite{hou2012image} provided another explanation from sparse representation, which states that, if the foreground is sparse in spatial domain and background is sparse in DCT domain (e.g., periodic textures), the spectral domain based approach will highlight the foreground region in the saliency map.

Motion, like color and texture, is also perceptually salient. \cite{huber2005visualizing} studied how three properties of motion, namely flicker, direction and velocity, contribute to this saliency. By setting the target object having different flicker rate, moving direction or motion velocity from the other objects, the target object can be easily identified by human subjects, i.e., being salient. In spectral, the target object and other objects can be mapped to two different bands (frequency and orientation), where the band corresponding to the target object has a much lower response than the band for the other object. Thus if we set the magnitude of the spectral to one, as the proposed method dose, the band for the other objects will be suppressed more than the target object, which makes the target object ``poped out" in the output. In Sec. \ref{sec:sim}, we will verify this analysis with experiments on synthetic data.
\subsection{Relationship to Existing Works}
Our method is related to some existing works and based on the way in which they computed the temporal information, we can roughly divide them into two categories:
\begin{enumerate}
\item Methods of the first category represent the temporal information by the motion, e.g., by frame differences \cite{ma2005generic} or by more dedicated motion estimation method including homography of adjacent frames \cite{zhai2006visual}  and phase correlation \cite{bian2009biological}. However, methods of this temporal information typically have limited temporal span, e.g., two adjacent frames (\cite{zhang2009sunday} tried to compute the frame differences of frames at a predefined sets of temporal spans), thus they are not sufficient for modeling the complex motion patterns.
\item In this category, the saliency of a spatiotemporal cuboid (refer as cuboid later) is measured by the ``differences" of this cuboid to other cuboids of the video or the template in the dictionaries, which may require high computational cost and/or require additional training data. The ``differences" of cuboids can be measured by distances \cite{seo2011static}, relative entropy \cite{li2010visual} mutual information \cite{mahadevan2010spatio} and coding length increments \cite{ban2008dynamic}.
\end{enumerate}
The proposed method is different from these methods. First, it does not rely on prior knowledge. Instead, it explores within the input video to detect the potential ``outliers". Second, the ``outliers" are found by exploring the whole temporal span. This makes the proposed algorithm be able to detect salient patterns from complex dynamic background. In addition, the propose method has low computational costs and is easy to implement. Fourier transform for multiple dimensional data can be computed as a sequence of 1D Fourier transform on each coordinate of the data, thus the computational cost of 3D Fourier transform for data $X\in\mathbb{R}^{M\times N\times T}$ is $O(MNT\mbox{log}(MNT))$. Thus the total computation cost for the proposed algorithm is $O(KMNT\mbox{log}(MNT))$, where $K$ is the number of feature channels.

\section{Experiment}\label{sec:experiment}
In this section, we evaluate the proposed method in saliency detection on both syntheic data (Sec. \ref{sec:sim}) and on two real image datasets (Sec. \ref{sec:diem}), CRCNS-ORIG and DIEM. The performance of the proposed methods are compared with the existing methods, some of which are state-of-art.
\subsection{Simulation Experiment}\label{sec:sim}
In this section, we evaluate the proposed method on synthetic data. In \cite{huber2005visualizing}, how three properties of motion, namely flicker, direction and velocity, contribute to the saliency was studied. In this section, we generate the synthetic data according to the their protocol. The input data is a short clip where the resolution is $174\times174$ with $400$ frames at the frame rate of $60$ frames per second. We put $36$ objects of size $5\times13$ in a $6\times6$ grid and a target object is randomly selected out of those $36$ objects. All the objects are allowed to move within a $29\times29$ region centered at their initial position (and warped back, if they move out of this region). The video is black-and-white. We design the following three experiments:
\begin{enumerate}
\item\textbf{Flicker}: we set the objects on-off at a specified rate and the target object at a different rate from the other $35$ objects;
\item\textbf{Direction}: we set the objects moving in a specified direction and the target object in a different direction. The velocity of all the objects are the same;
\item\textbf{Velocity}: we set the objects moving in a specified velocity and the target object moves in a different velocity. The moving direction of the all the objects are the same.
\end{enumerate}
All the other parameters are the same as used in \cite{huber2005visualizing}. According to \cite{huber2005visualizing}, the target object could be easily identified by human subjects, when its motion property (e.g., flicker rate, moving direction, velocity) is different from the other objects. We also include some ``blind" trials, where the target object has the same motion property as the other $35$ objects. In this case, the target object can't be identified by the human subjects, i.e., there is no salient region.

\begin{figure*}[!ht]
  \centerline{\includegraphics[width=16cm]{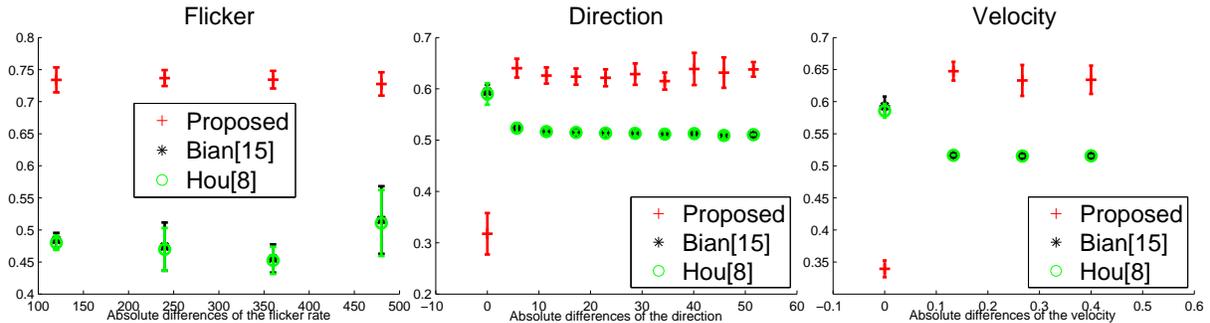}}
   \caption{The AUC on the synthetic data for the proposed method and two existing methods. For ``Direction" and ``Velocity", we also include some ``blind" trials (X-axis has value $0$), where the target object has exactly the same motion property as the other $35$ objects. In those trials, the target object can't be identified by human subjects, i.e., there is no salient object \cite{huber2005visualizing}.}\label{fig:synthetic}
\end{figure*}

We apply the proposed method to the input data. For comparison, we also evaluate the method proposed in \cite{bian2009biological} and \cite{hou2012image}. We use the area under receiver operating characteristic curve as the performance metric. The ground truth mask is generated according to the location of the target object. The experiment result is shown in \ref{fig:synthetic}.
\begin{figure*}[!ht]
\begin{minipage}[b]{0.24\linewidth}
  \centering
  \centerline{\includegraphics[width=4cm]{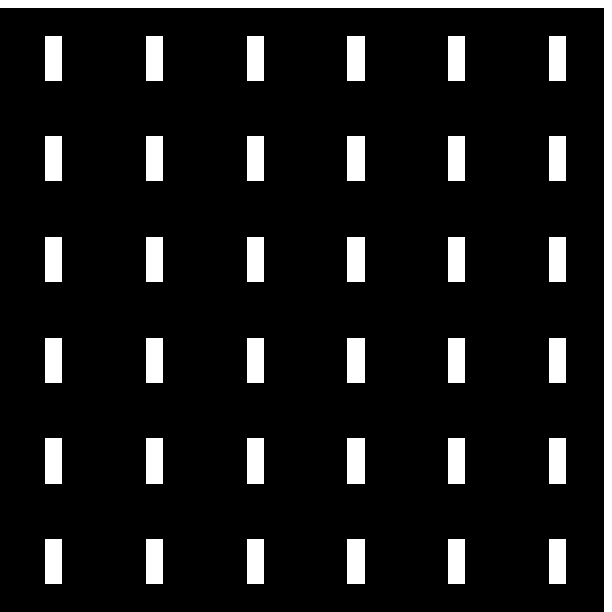}}
  \centering{Blind}
\end{minipage}
\hfill
\begin{minipage}[b]{0.24\linewidth}
  \centering
  \centerline{\includegraphics[width=4cm]{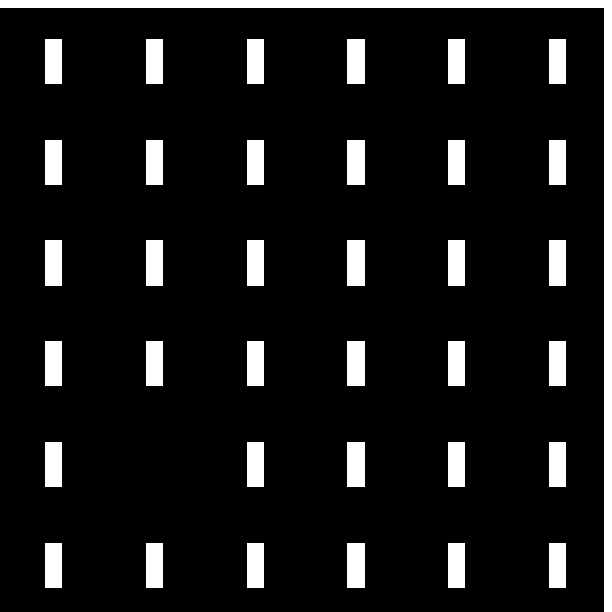}}
  \centering{Flicker}
\end{minipage}
\hfill
\begin{minipage}[b]{0.24\linewidth}
  \centering
  \centerline{\includegraphics[width=4cm]{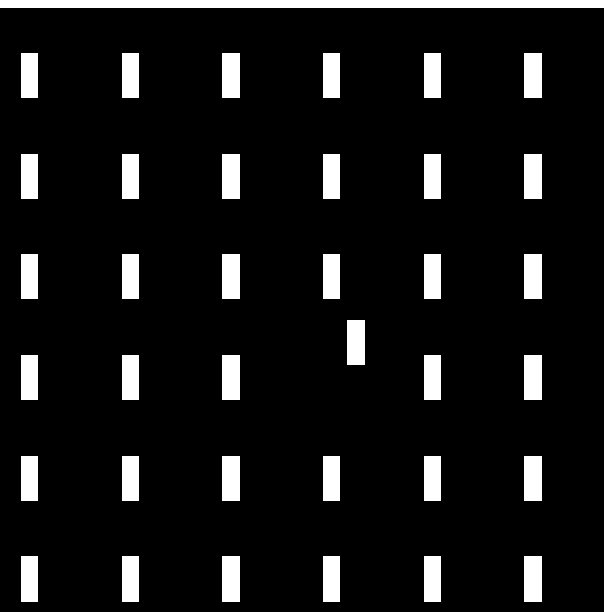}}
  \centering{Direction}
\end{minipage}
\hfill
\begin{minipage}[b]{0.24\linewidth}
  \centering
  \centerline{\includegraphics[width=4cm]{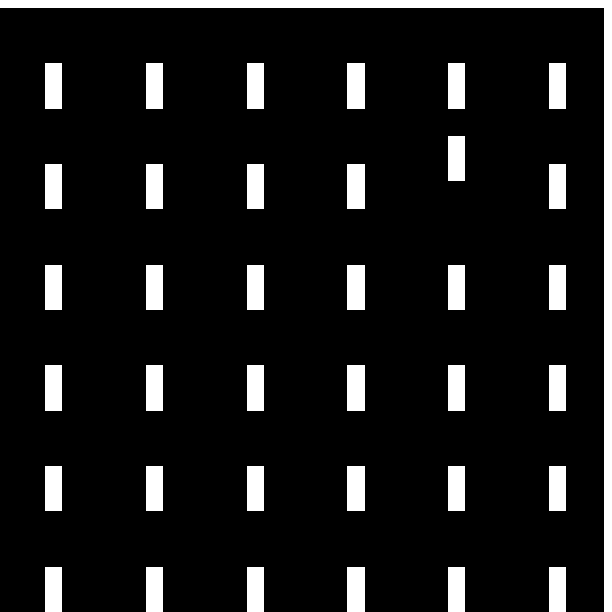}}
  \centering{Velocity}
\end{minipage}
   \caption{Some visual sample of the synthetic data for different experiments.}\label{fig:synthetic_sample}
\end{figure*}

From the experiment results, we can find that the proposed method detects the salient region much more accurately than \cite{bian2009biological} and \cite{hou2012image} in all except the ``blind" trials. For the ``blind" trials, the AUC for the proposed method significantly reduces, which shows that the proposed method is also robust. However, \cite{bian2009biological} and \cite{hou2012image} don't survive in those ``blind" trials. Surprisingly, \cite{bian2009biological} and \cite{hou2012image} achieves quite similar performances, though \cite{bian2009biological} was supposed to achieve better result as it include the differences of two adjacent frames as motion (temporal) information.
\subsection{Spatiotemporal Saliency Detection}\label{sec:diem}
In previous section, we test the proposed spatiotemporal saliency detector on synthetic videos, with the comparison to two other saliency detectors, where the proposed detector shows better performances in capturing the temporal information. In this section, we evaluate the proposed spatiotemporal saliency detector on two challenging video datasets for saliency evaluation, CRCNS-ORIG \cite{itti2009crcns} and DIEM \cite{mital2011clustering}. For this experiment, we first convert each frame into the LAB color space, then compute the spatiotemporal saliency in each channel independently and the final spatiotemporal saliency is the summation of the saliency maps of all three channels.

\begin{figure}[!ht]
  \centerline{\includegraphics[width=8cm]{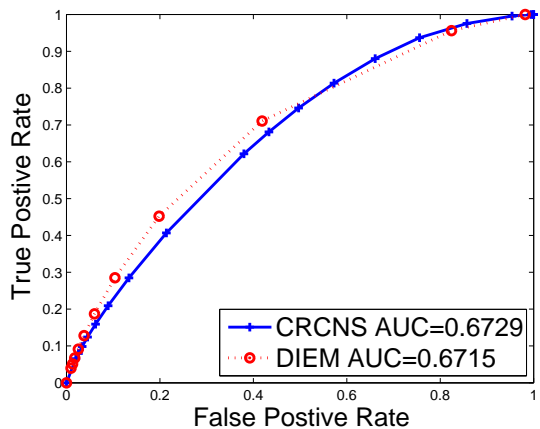}}
   \caption{The receiver operating characteristic curve of the propose method in CRCNS-ORIG dataset and DIEM dataset. The area under the curve is $0.6639$ and $0.6896$ accordingly.}\label{fig:crcns_diem}
\end{figure}

CRCNS-ORIG includes 50 video clips from different genres, including TV programs, outdoor scenes and video games. Each clip is 6-second to 90-second long at 30 frames per second. The eye fixation data is captured from eight subjects with normal or correct-normal vision. In our experiment, we downsample the video from $640\times480$ to $160\times120$ and keep the frame rate untouched, then apply the our spatiotemporal saliency detector. To measure the performance, we compute the area under curve (AUC) and F-measure (harmonic mean of true positive rate and false positive rate). The experiment result is shown in Fig. \ref{fig:crcns_diem}, where the area under curve (AUC) is $0.6639$ and F-measure is $0.1926$. Tab. \ref{tab:crcns_diem} compares the result of the proposed method with some state-of-art methods on CRCNS-ORIG, which indicates that our method outperforms them by at least $0.06$ regarding AUC. The per-video AUC score is shown in Fig. \ref{fig:crcns} in Appendix \ref{sec:auc}.
\begin{table}
\begin{center}
\begin{tabular}{|c|c||c|c|}
\hline
Method & AUC  & Method & AUC\\\hline
AWS \cite{garcia2009decorrelation} & $0.6000$&AWS \cite{garcia2009decorrelation} & $0.5770$\\\hline
HouNIPS \cite{hou2008dynamic} & $0.5967$&Bian \cite{bian2009biological} & $0.5730$\\\hline
Bian \cite{bian2009biological} & $0.5950$&Marat \cite{marat2009modelling} & $0.5730$\\\hline
IO & $0.5950$&Judd \cite{judd2009learning} & $0.5700$\\\hline
SR \cite{hou2007saliency} & $0.5867$&AIM \cite{bruce2005saliency} & $0.5680$\\\hline
Torralba \cite{torralba2003modeling} & $0.5833$&HouNIPS \cite{hou2008dynamic} & $0.5630$\\\hline
Judd \cite{judd2009learning} & $0.5833$&Torralba \cite{torralba2003modeling} & $0.5840$\\\hline
Marat \cite{marat2009modelling} & $0.5833$&GBVS \cite{harel2006graph} & $0.5620$\\\hline
Rarity-G \cite{mancas2007computational} & $0.5767$&SR \cite{hou2007saliency} &$0.5610$ \\\hline
CIOFM \cite{itti2006bayesian} & $0.5767$&CIO \cite{itti2006bayesian} & $0.5560$\\\hline\hline
Proposed & \textbf{$0.6639$}&Proposed & \textbf{$0.6896$}\\\hline
\end{tabular}
\end{center}
\caption{The result the proposed method compared with the results of the top ten existing methods on CRCNS dataset (left) and DIEM dataset (right) according to \cite{borji2012quantitative}. From this table, we can find that the propose method gets obvious better performances than the state-of-arts on both two datasets.}\label{tab:crcns_diem}\vspace{-5mm}
\end{table}

DIEM dataset collects data of where people look during dynamic scene viewing such as film trailers, music videos, or advertisements. It currently consists of data from over $250$ participants watching $85$ different videos. Each video in DIEM dataset includes $1000$ to $6000$ frames at $30$ frames per second. Similarly as CRCNS, we downsample the video to $1/4$ (e.g., from $1280\times720$ to $320\times180$) while maintaining the aspect ratio and frame rate. We observe that each video in DIEM dataset is consisted of a sequence of short clips, where each clip has $30$ to $100$ frames. To properly detect the saliency from those videos, we apply the window function to our spatiotemporal saliency detector, where the size of the window (along temporal direction) is $60$-frame. The experiment result is shown in Fig. \ref{fig:crcns_diem} and Tab. \ref{tab:crcns_diem}, where the AUC is $0.6896$ and F-measure is $0.35$. From the table, we can find that the proposed method outperforms the state-of-arts by over $10\%$. The per-video AUC score is shown in Fig. \ref{fig:diem} in Appendix \ref{sec:auc}.
\section{Application of Spatiotemporal Saliency}\label{sec:application}
In the previous section, we show that the proposed method is able to detect the saliency region in the video. The saliency detection for image has been used more and more in other visual tasks, e.g., image segmentation, object recognition. A natural question arises that can we also appliy the spatiotemporal saliency detection for some important vision tasks. In this section, we show how can we applied the spatiotemporal saliency computed by the proposed methods to two important vision tasks, i.e., abnormality detection (\ref{sec:abnormal}) and spatiotemporal interest pointer detection (\ref{sec:stip}).
\subsection{Abnormality Detection}\label{sec:abnormal}
According to our previous analysis, the salient region should be different from the neighbor, both spatially and temporally. This spatiotemporal saliency shares a lot of common to the concept of abnormality in video. Thus in this section, we show how can we utilize the proposed spatiotemporal saliency detector to detect abnormality from the video.

For abnormality detection, we start with computing the saliency map for the input video as described above. The regions containing abnormalities can be detected by founding the region where the saliency value is above a threshold. Then the saliency score of a frame is computed as the average of saliency value of the pixels in that frame, i.e.,
\begin{equation}
\textbf{s}(t)=\frac{1}{NM}\sum_i{\sum_j{\textbf{X}(i,j,t)}}
\end{equation}
where $\textbf{s}(t)$ is the saliency score of $t_{th}$ frame, $N\times M$ is the size of one frame, $i$, $j$, $t$ are row, column and frame index of the 3D saliency map accordingly. The frame with high saliency score would contain abnormality.

We evaluate the proposed method for abnormality detection in videos from two datasets: UMN abnormal dataset\footnote{\url{http://mha.cs.umn.edu/Movies/Crowd-Activity-All.avi}} and  UCSD dataset \cite{mahadevan2010anomaly}. Abnormal detection has attracted a lot efforts from the researchers. However, most of the existing works require training stage, e.g., social force \cite{mehran2009abnormal}, sparse reconstruction \cite{cong2011sparse}, MPPCA \cite{kim2009observe}, i.e., they need training data to initialize the model. The proposed method, instead, dose \textbf{NOT} need any training stage or training data.

\begin{figure}[!ht]
  \centerline{\includegraphics[width=8cm]{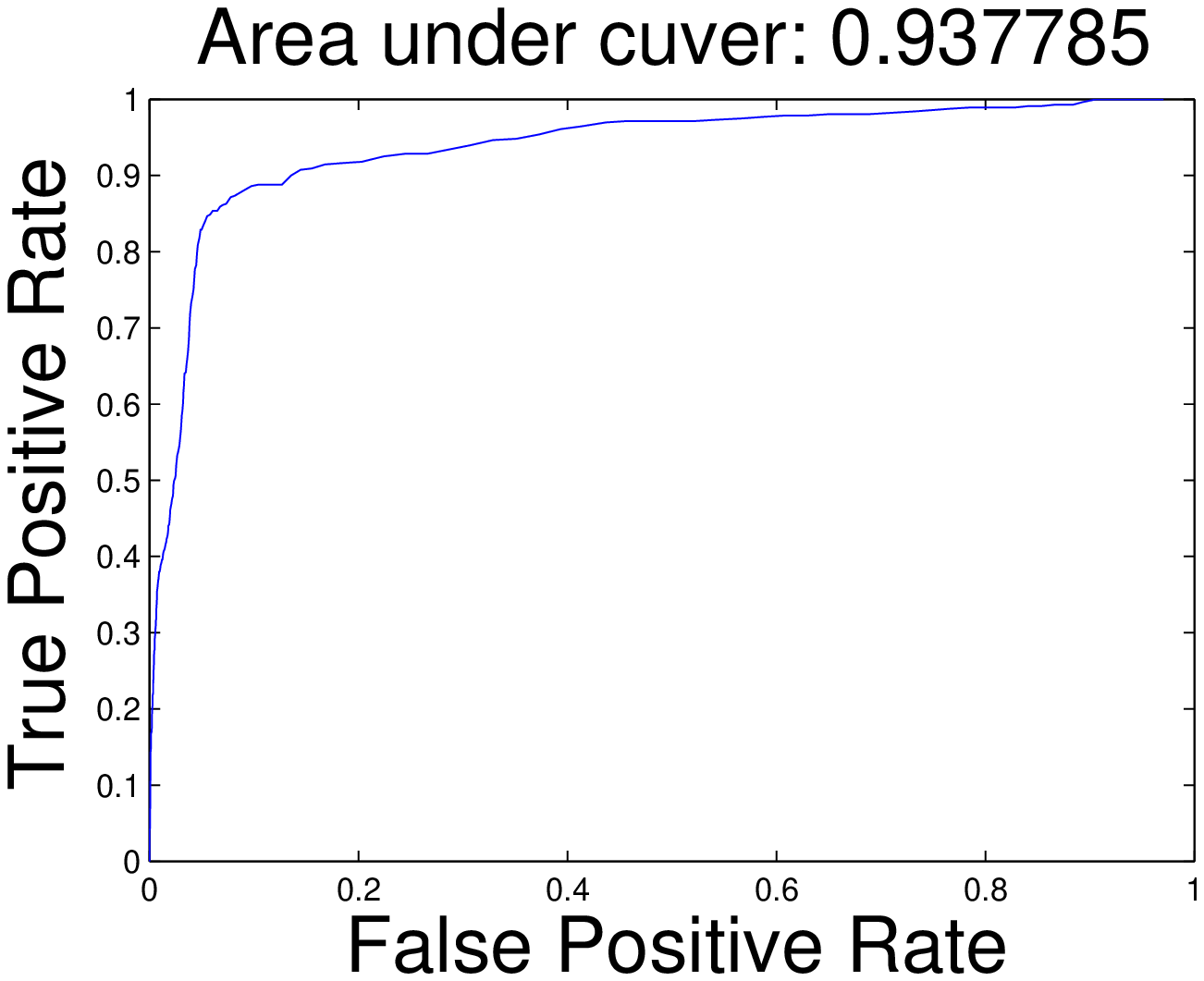}}
   \caption{The ROC for the UMN dataset computed with the propose method.}\label{fig:umn_result}
\end{figure}

\begin{figure}[!ht]
\begin{minipage}[b]{1\linewidth}
  \centering
  \centerline{\includegraphics[width=8cm]{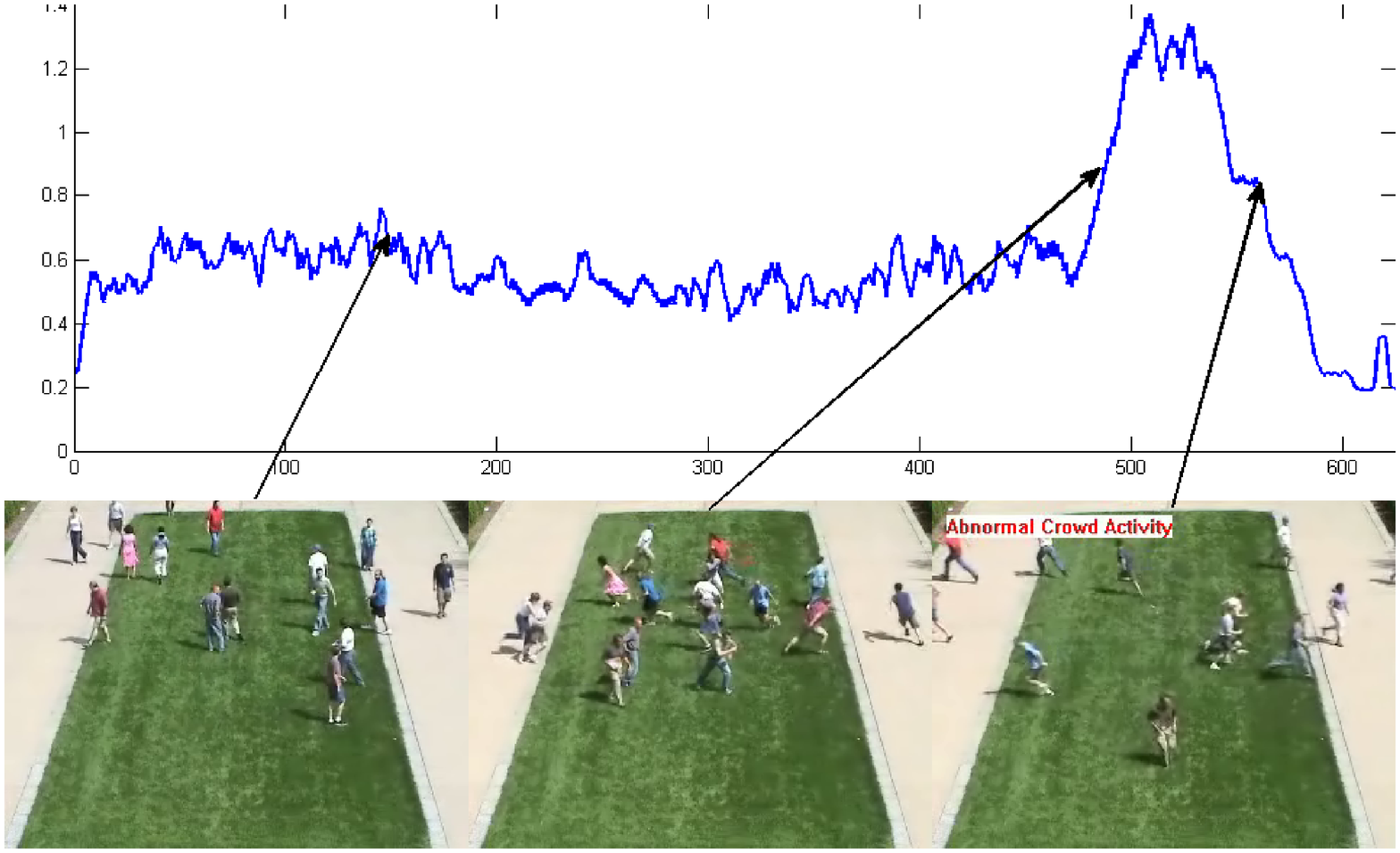}}
  \centering{Scene 1}
\end{minipage}
\vfill
\begin{minipage}[b]{1\linewidth}
  \centering
  \centerline{\includegraphics[width=8cm]{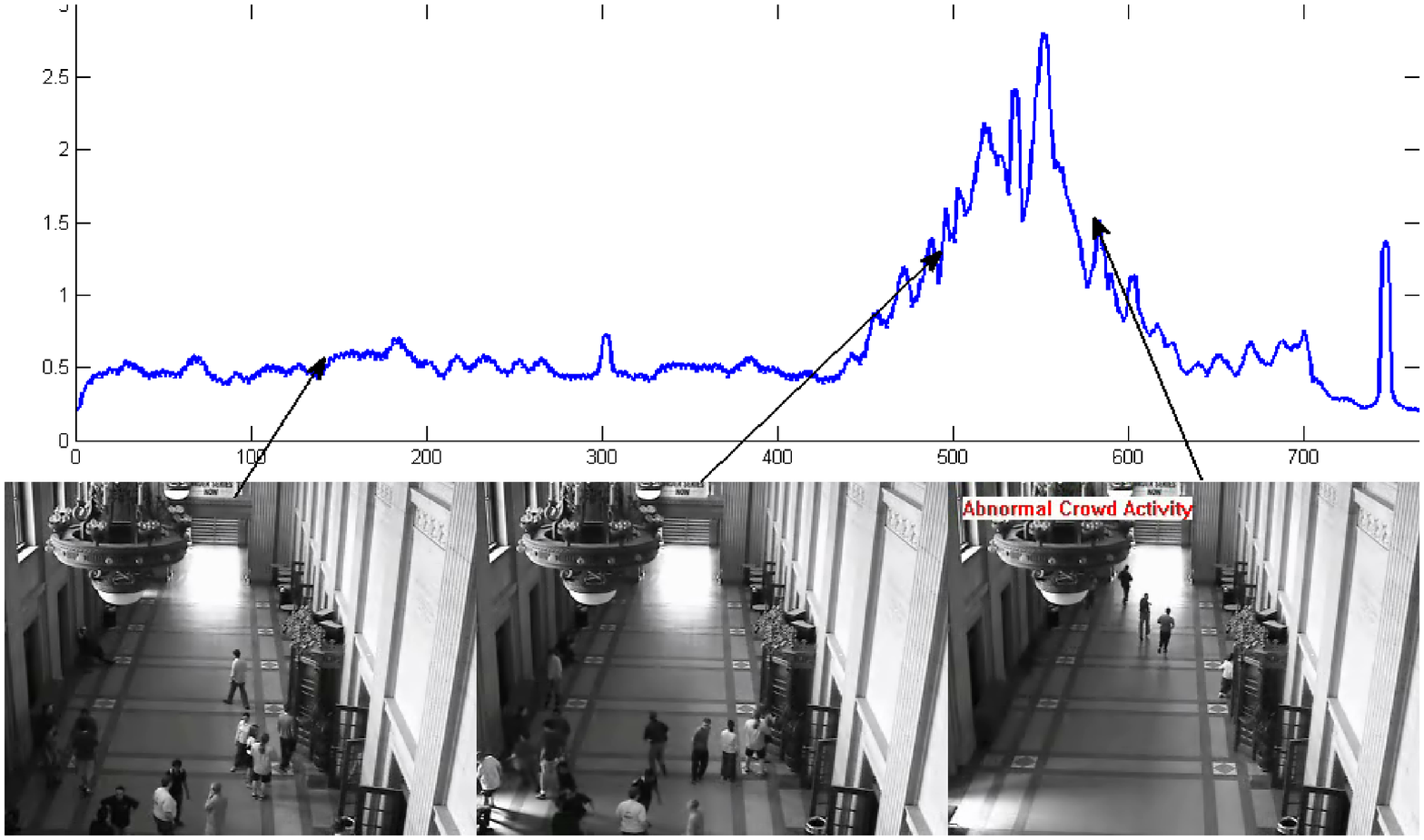}}
  \centering{Scene 2}
\end{minipage}
\vfill
\begin{minipage}[b]{1\linewidth}
  \centering
  \centerline{\includegraphics[width=8cm]{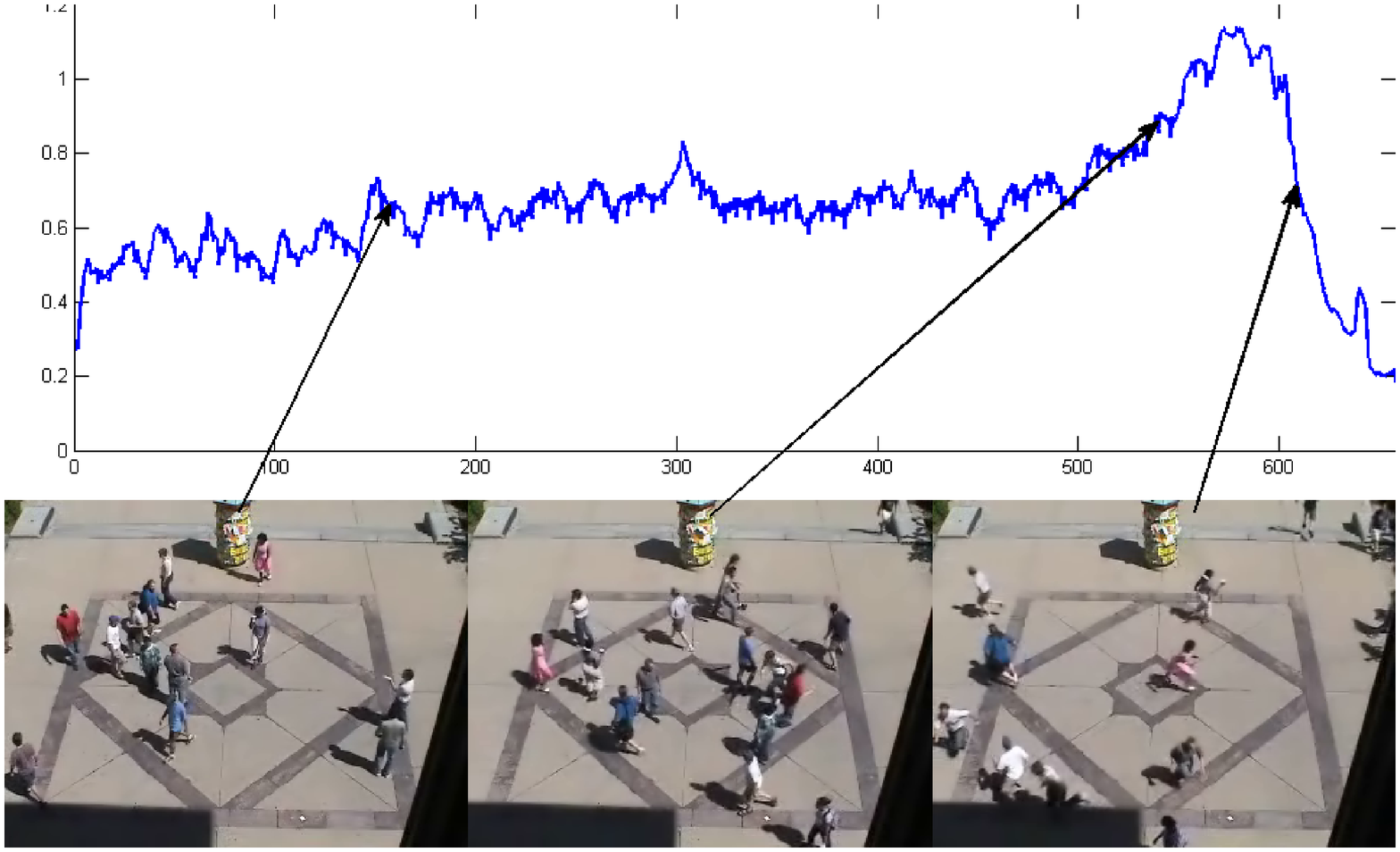}}
  \centering{Scene 3}
\end{minipage}
   \caption{Some sample results for the UMN datasets, where we pick one video for each scene. The top is the saliency value (Y-axis) for each frame (X-axis) and bottom are sample frames picked from different frames (as shown by the arrow).}\label{fig:umn}\vspace{-5mm}
\end{figure}

\begin{figure}[!ht]
  \centerline{\includegraphics[width=8cm]{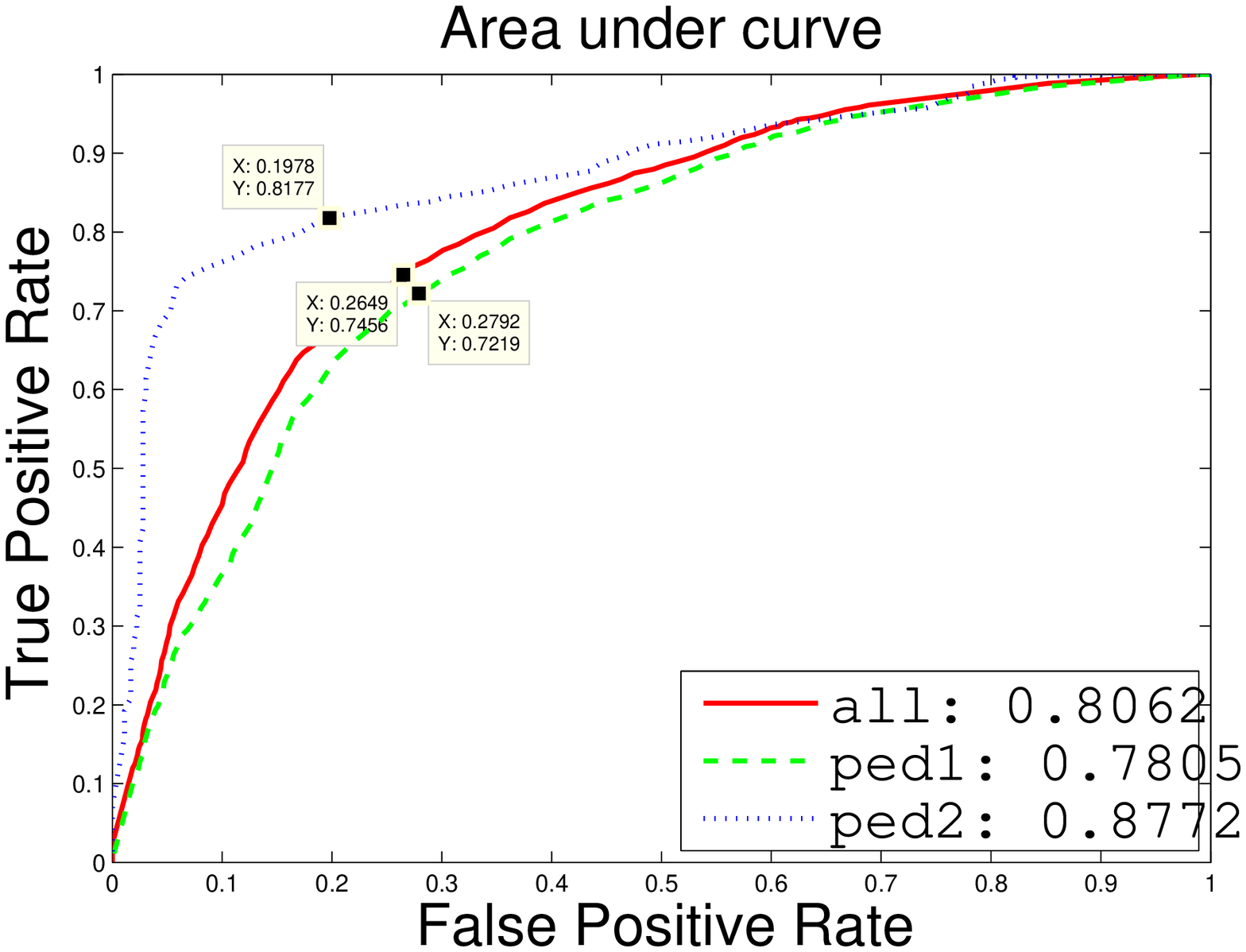}}
   \caption{The ROC for the UCSD dataset computed with the propose method.}\label{fig:ucsd_auc}\vspace{-3mm}
\end{figure}

\begin{figure}[!ht]
\begin{minipage}[b]{0.4\linewidth}
  \centering
  \centerline{\includegraphics[width=4cm]{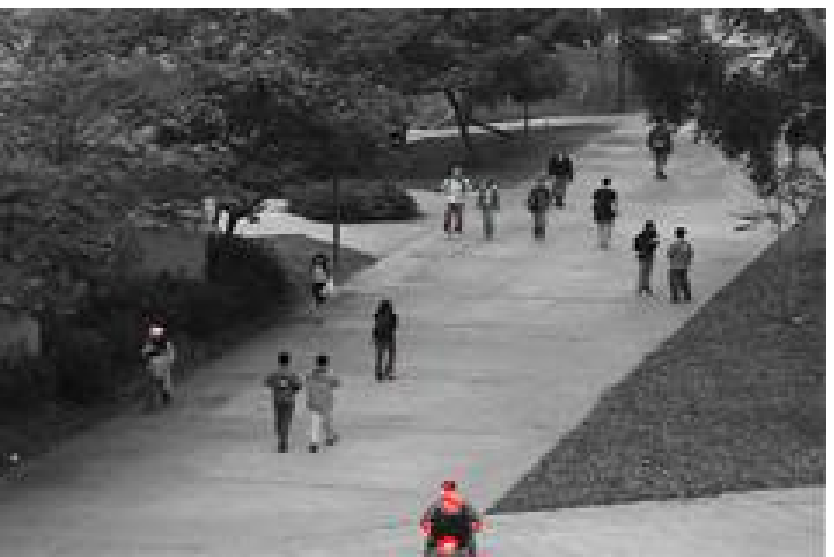}}
  \centering{Peds1: Wheelchair}
\end{minipage}
\hfill
\begin{minipage}[b]{0.4\linewidth}
  \centering
  \centerline{\includegraphics[width=4cm]{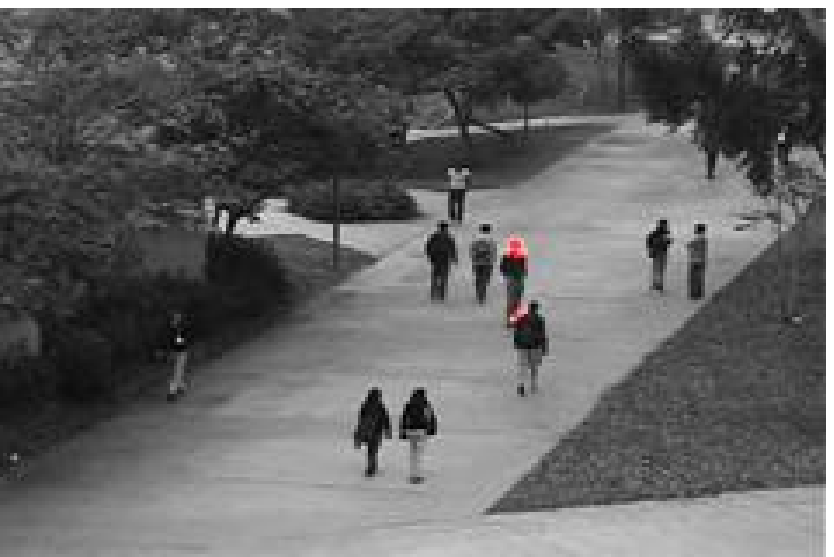}}
  \centering{Peds1: Skater}
\end{minipage}
\vfill
\begin{minipage}[b]{0.4\linewidth}
  \centering
  \centerline{\includegraphics[width=4cm]{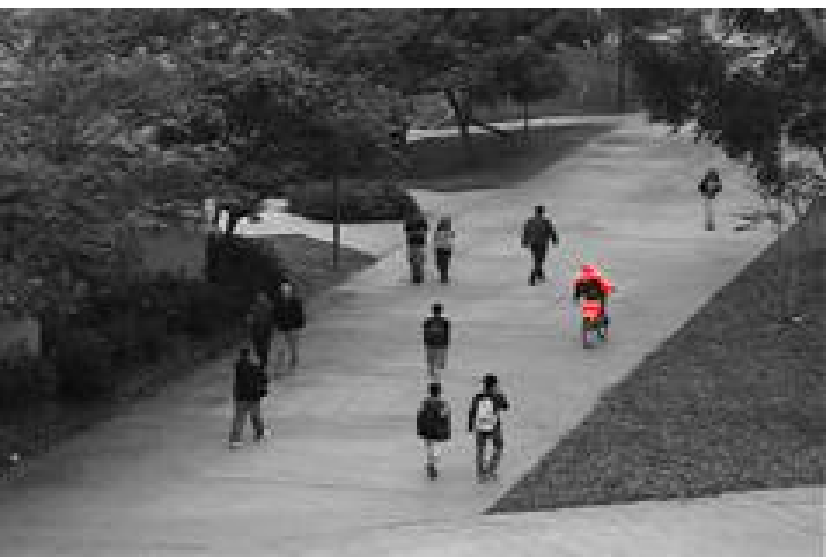}}
  \centering{Peds1: Bike}
\end{minipage}
\hfill
\begin{minipage}[b]{0.4\linewidth}
  \centering
  \centerline{\includegraphics[width=4cm]{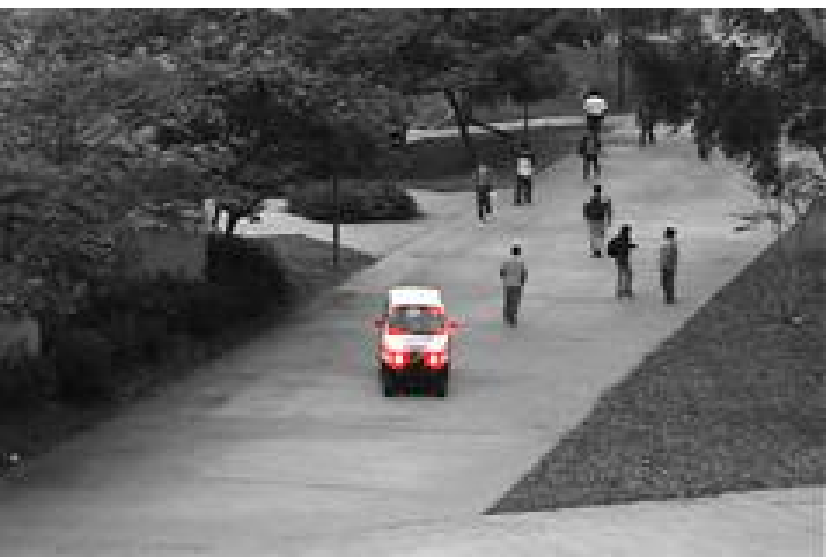}}
  \centering{Peds1: Cart}
\end{minipage}
\vfill
\begin{minipage}[b]{0.4\linewidth}
  \centering
  \centerline{\includegraphics[width=4cm]{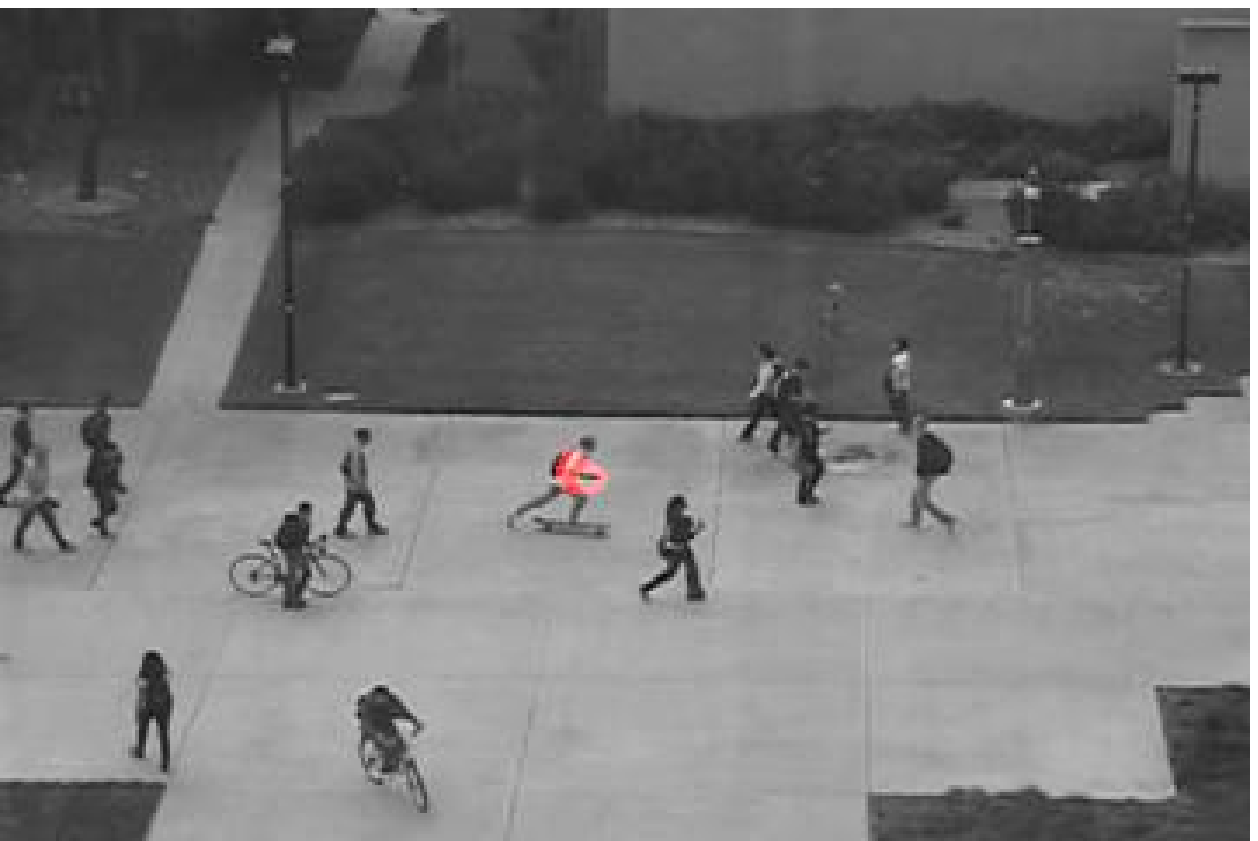}}
  \centering{Peds2: Skater}
\end{minipage}
\hfill
\begin{minipage}[b]{0.4\linewidth}
  \centering
  \centerline{\includegraphics[width=4cm]{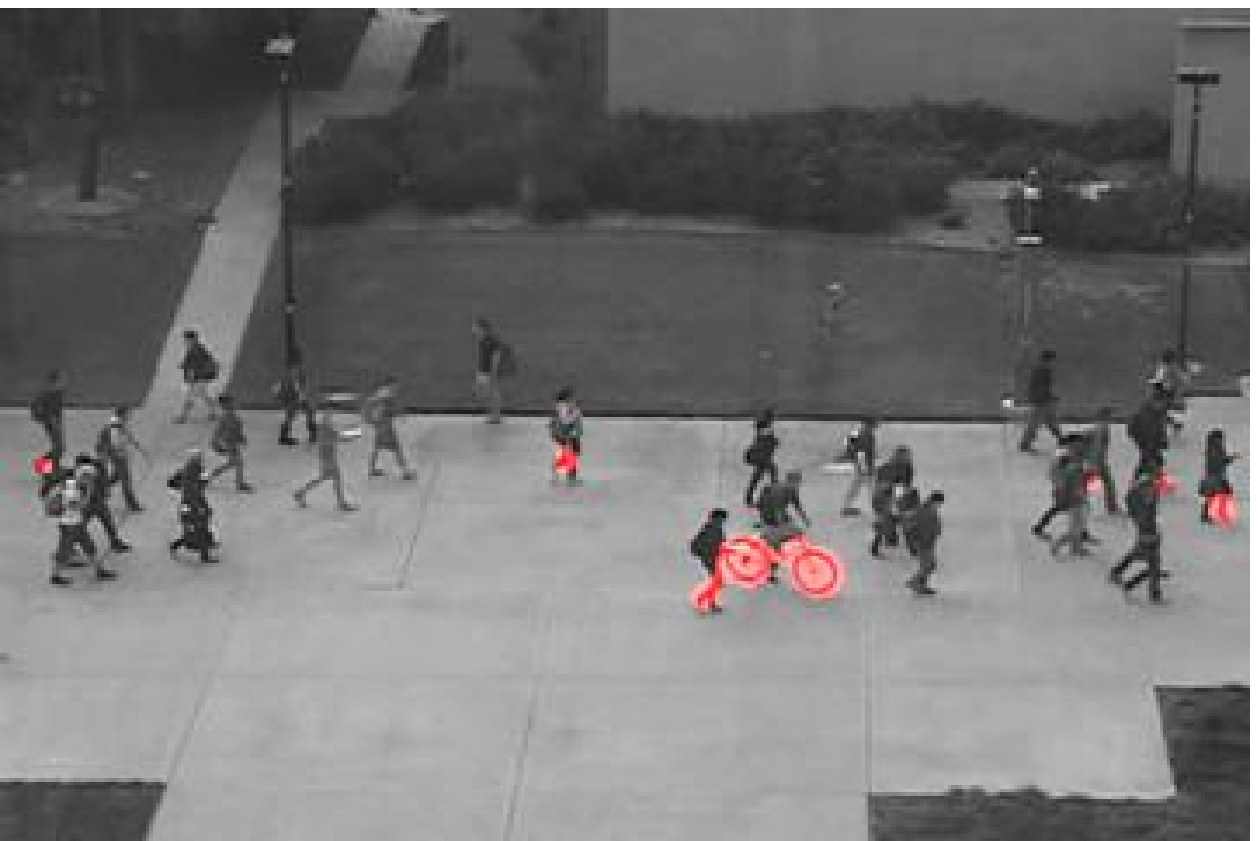}}
  \centering{Peds2: Bike}
\end{minipage}
   \caption{Some sample results for the UCSD datasets, where the red color highlights the detected abnormal region, i.e., the saliency value of the pixel is higher than four times of the mean saliency value of the video.}\label{fig:ucsd}\vspace{-3mm}
\end{figure}

The result on UMN abnormal dataset is shown in Tab. \ref{tab:umn}, where we compute the frame-level true positive rate and false positive rate then compute the area under the ROC (Fig. \ref{fig:umn_result}). Fig. \ref{fig:umn} shows the result for videos of three scenes, where we plot saliency value of each frame and show some sample frames. The result on UCSD dataset is shown in Tab. \ref{tab:ucsd}, where we report frame-level equal-error rate (EER) \cite{mahadevan2010anomaly}. Fig. \ref{fig:ucsd_auc} shows the ROC for UCSD dataset with the proposed method; Fig. \ref{fig:ucsd} shows eight samples frames, where red color highlights abnormal regions. We can find that, without training data, the proposed method still outperforms several state-of-arts in the literature, e.g., social force, MPPCA.
\begin{table}
\begin{center}
\begin{tabular}{|c|c|}
\hline
Method & AUC\\\hline
Optical flow \cite{mehran2009abnormal} & $0.84$\\\hline
Social force \cite{mehran2009abnormal} & $0.96$\\\hline
Chaotic invariants \cite{wu2010chaotic} & $0.99$\\\hline
NN \cite{cong2011sparse} & $0.93$\\\hline
Sparse reconstruction \cite{cong2011sparse} & $0.978$\\\hline
Interaction force \cite{raghavendra2011optimizing} & $0.9961$\\\hline
Proposed & $0.9378$\\\hline
\end{tabular}
\end{center}
\caption{The result on UMN dataset. Note, we have cropped out the region which contains the text ``abnormal", and results in frame resolution $214\times320$. Please note that, most of those methods, except the proposed one, need a training stage.}\label{tab:umn}\vspace{-5mm}
\end{table}
\
\begin{table}
\begin{center}
\begin{tabular}{|c|c|c|c|}
\hline
Method & Ped1 & Ped2 & Overall\\\hline
Social force \cite{mehran2009abnormal} & $31\%$ & $42\%$& $37\%$\\\hline
MPPCA \cite{kim2009observe}& $40\%$& $30\%$& $35\%$\\\hline
MDT \cite{mahadevan2010anomaly}& $25\%$&$25\%$&$25\%$\\\hline
Adam \cite{adam2008robust}&$38\%$&$42\%$&$40\%$\\\hline
Reddy \cite{reddy2011improved}&$22.5\%$&$20\%$&$21.25\%$\\\hline
Sparse \cite{cong2011sparse}&$19\%$&$N.A.$&$N.A.$\\\hline
Proposed & $27\%$ & $19\%$ & $23\%$\\\hline
\end{tabular}
\end{center}
\caption{The frame level EER (the lower the better) for UCSD dataset. Please note that, most of those methods, except the proposed one, need a training stage. From the result, we can found that the proposed method, even without traing stage or training data, can still outperform social force, MPPCA.}\label{tab:ucsd}\vspace{-3mm}
\end{table}
\subsection{Spatiotemporal Saliency Point Detector}\label{sec:stip}
The regions which attracts human's attention most would contribute most to people's perception of the scene. The saliency map computed with the proposed method will hightlight those regions. Thus we propose to sample the interest points based on the saliency map of the data, which we refer as spatiotemporal saliency point detector (STSP).

To detect interest point, we also starts with computing the saliency map $\textbf{Z}$ for the input data $\textbf{X}$. Then we apply non-maximum suppression on the saliency map to sample the interest points: an interest point is selected at $(x,y,t)$ if and only if
\begin{eqnarray}
\textbf{Z}(x,y,t)&\geq &\rho\\\nonumber
\textbf{Z}(x,y,t)&\geq &\textbf{Z}(i,j,k)\ \forall(i,j,k)\in N(x,y,t)
\end{eqnarray}
where $\rho$ is a predefined threshold (e.g., $2\mu$) and $N(x,y,t)$ is the set of positions near $(x,y,t)$.

Similar as \cite{laptev2005space}, for each interest points $(x,y,t)$, we extract a descriptor within its neighbor area characterized as $(x,y,t,\sigma,\tau)$, where $(x,y,t)$ is the center, $\sigma$, $\tau$ are the spatial and temporal scales (we use $18\times18\times10$, $25\times25\times14$ and $36\times36\times20$ here) accordingly. The neighbor is further divided into multiple subblocks (e.g., $3\times3\times2$ along spatial and temporal direction accordingly); for each subblock, we computed the 3D gradient $g=[g_x,g_y,g_t]$; then we quantize the orientations of the gradients into a histogram of four bins; finally the histogram of each subblock is normalized to unit $l_1$ norm and histograms of all subblocks is concatenated into one histogram, i.e., the descriptor for interest point $(x,y,t)$.

Compared with existing spatiotemporal interst point detectors, which mostly choose the location where the gradient is strong and stable cross different scales. However, the gradient is a low level information and is insufficient to capture the complex dynamics as the human vision does. Instead, the proposed method explores the relationship of each location over all spatial and temporal span, which is able to model complex dynamics in the video.

For evaluation, we use three datasets: Weizmann dataset \cite{lena2007actions}, KTH dataset \cite{schuldt2004recognizing} and UCF sports dataset \cite{rodriguez2008action}. Since the method is proposed for detecting interest points, we only compare it with several state-of-art spatiotemporal interest point detectors including Harris3D \cite{laptev2005space}, Gabor \cite{dollar2005behavior}, Hessian3D \cite{willems2008efficient} and dense sampling \cite{kla2010learning}, where the result are summarized in \cite{kla2010learning}. The parameters of the detectors are set as suggested by the paper accordingly.

\begin{figure}[!ht]
\begin{minipage}[b]{0.48\linewidth}
  \centering
  \centerline{\includegraphics[width=4cm]{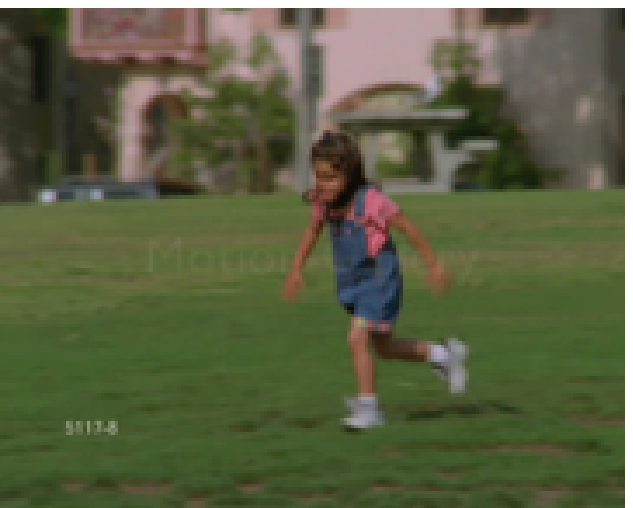}}
\end{minipage}
\hfill
\begin{minipage}[b]{0.48\linewidth}
  \centering
  \centerline{\includegraphics[width=4cm]{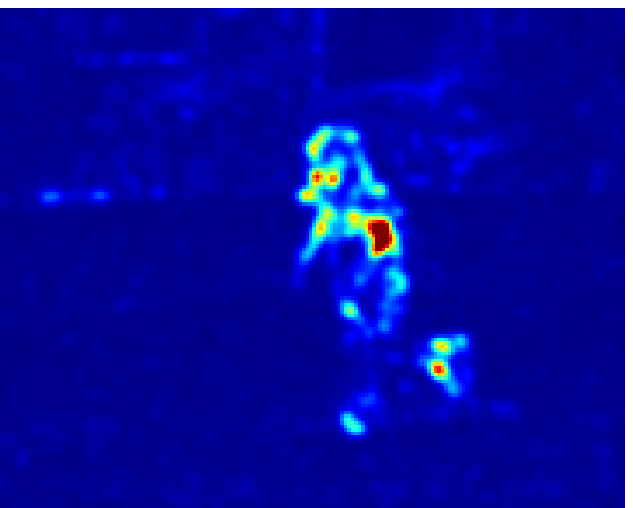}}
\end{minipage}
\vfill
\begin{minipage}[b]{0.48\linewidth}
  \centering
  \centerline{\includegraphics[width=4cm]{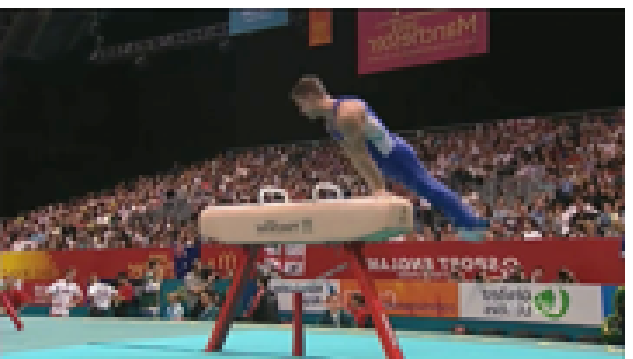}}
\end{minipage}
\hfill
\begin{minipage}[b]{0.48\linewidth}
  \centering
  \centerline{\includegraphics[width=4cm]{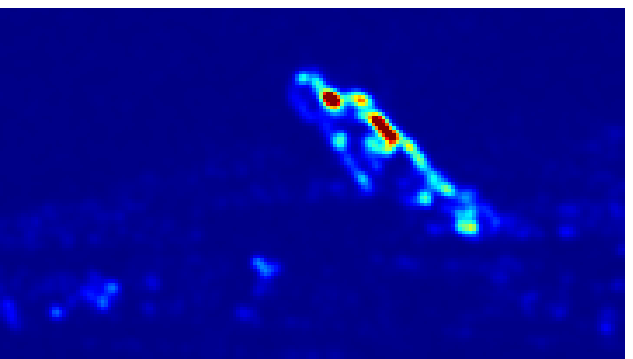}}
\end{minipage}
\vfill
\begin{minipage}[b]{0.48\linewidth}
  \centering
  \centerline{\includegraphics[width=4cm]{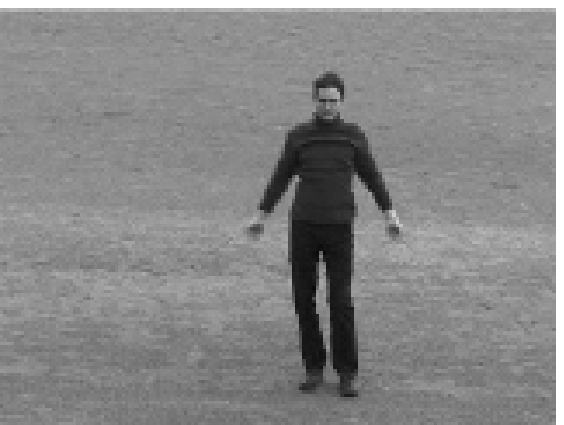}}
\end{minipage}
\hfill
\begin{minipage}[b]{0.48\linewidth}
  \centering
  \centerline{\includegraphics[width=4cm]{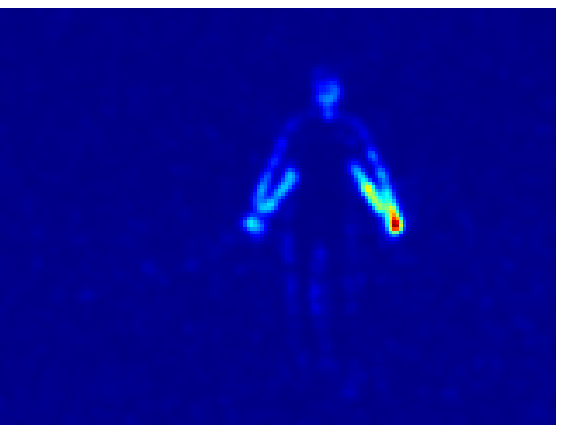}}
\end{minipage}
\vfill
\begin{minipage}[b]{0.48\linewidth}
  \centering
  \centerline{\includegraphics[width=4cm]{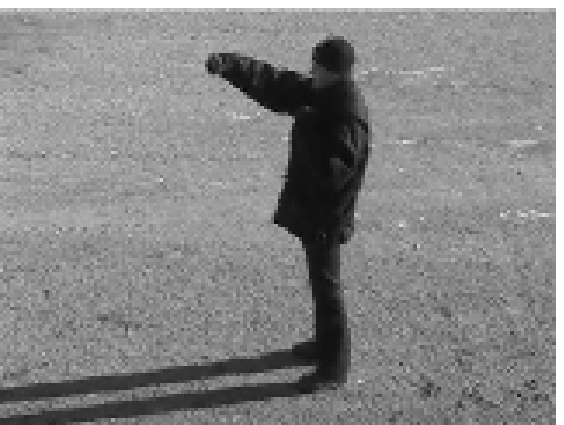}}
\end{minipage}
\hfill
\begin{minipage}[b]{0.48\linewidth}
  \centering
  \centerline{\includegraphics[width=4cm]{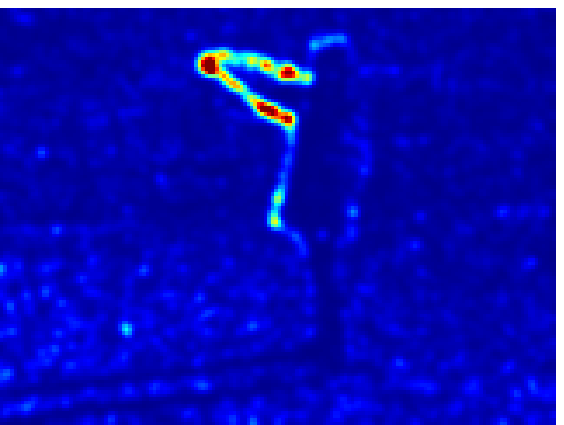}}
\end{minipage}
   \caption{Some samples frames (left) from UCF sports action dataset (Row 1, 2) and KTH dataset (Row 3, 4) with their saliency maps (right).}\label{fig:action}\vspace{-5mm}
\end{figure}

Fig. \ref{fig:action} shows the saliency map for some sample frames of videos from UCF sports action dataset and KTH dataset. From the figure, we can found that the saliency map computed with the proposed method highlights the moving region while suppressing the background. The proposed method is also robust to moving background (Row 1), clutter background (Row 2) and scale variation (Row 3). In addition, from Row 3 to 4, we can found the moving parts of body, e.g., hands, get higher saliency value (red color) then other body parts. The spatiotemporal saliency interest points will be mostly sampled from those hightlighted regions and augment the description of the action of interest.

To quantitatively evaluate the performances of different detector, we use the interest points detected by those detector to train a classifiers for activity recognition. We use both histogram of gradient (HoG) and histogram of optical flow (HoF) as the descriptor. Bag of words is used to represent the video, where each input is represented as a histogram of words in the codebook (size of of codebook is $k=2000$); then classifier (support vector machine with $\chi^2$ kernel) is applied to those histograms to classify the input. For Weizmann dataset and UCF sports dataset, we use leave-one-out scheme for training and testing; for KTH dataset, we follow the standard partition in \cite{schuldt2004recognizing}.

Tab. \ref{tab:action} reports the performances of different detectors on three dataset, where we test extracting feature on the original video and also extracting feature on the saliency map of the original video (refer as ``proposed*"). From the table we find that, the proposed method (and ``proposed*") achieves the best result over all three datasets. Especially ``proposed*" achieved the best results for KTH dataset and Weizmann data; ``proposed" achieved the best results for UCF sports action dataset. For video with simple background(e.g., KTH dataset and Weizmann dataset), extracting descriptor on saliency map instead of the video itself could be a better choice. 
\begin{table}
\begin{center}
\begin{tabular}{|c|c|c|c|}
\hline
Method & Weizmann & KTH & UCF sports\\\hline
Harris3D & $85.6\%$ & $91.8\%$& $78.1\%$\\\hline
Gabor& N.A.& $88.7\%$& $77.7\%$\\\hline
Hessian3D& N.A.&$88.7\%$&$79.3\%$\\\hline
Dense&N.A.&$86.1\%$&$81.6\%$\\\hline
Proposed&$84.5\%$&$88.0\%$&$\textbf{86.7}\%$\\\hline
Proposed*&$\textbf{95.6}\%$&$\textbf{92.6}\%$&$85.6\%$\\\hline
\end{tabular}
\end{center}
\caption{The performances of different detectors on three datasets. For ``proposed*", we extract the descriptor on the saliency map instead of on the video.}\label{tab:action}\vspace{-5mm}
\end{table}
\section{Conclusion and Discussion}\label{sec:conclusion}
In this paper, we proposed a novel approach for detecting spatiotemporal saliency, which was simple to implement and computationally efficient. The proposed approach was inspired by recent development of spectrum analysis based visual saliency approaches, where phase information was used for constructing the saliency map of the image. Recognizing that the computed saliency map captured the region of human's attention for dynamic scenes, we proposed two algorithms utilizing this saliency map for two important vision tasks. These approaches were evaluated on several well-known datasets with comparisons to the state-of-arts in the literature, where good results were demonstrated. For the future work, we will focus on theoretical analysis of the proposed method and the analysis on the selection of the window function.


%

\appendix
\section{Appendix}\label{sec:qft}
To compare the performances of combining four visual cues via QFT and performances via summation of saliency maps of each visual cues, we design the following experiment. We run $1000$ simulations and in each simulation we generate a $r\times c\times 4$ array, where r and c is a random number between $[1,1000]$  and $4$ is the number of feature channels. We compute the saliency map with different methods then measures their similarities via cross-correlation, where 0.91 is reported for QFT and FFT. After smoothing the saliency map with a Gaussian kernel, the correlation is over 0.998. For natural image, we could expect an even higher correlation. 

This suggests that, we can compute the saliency map for each visual cue independently and then add them together, which will yield quite similar result by using quaternion Fourier transform. In addition, the proposed method other than QFT provides more flexibility, e.g., we can assign different weights to the visual cues as \cite{judd2009learning}.
\subsection{Supplementary Results}\label{sec:auc}
\begin{figure}[h]
  \centerline{\includegraphics[width=16cm]{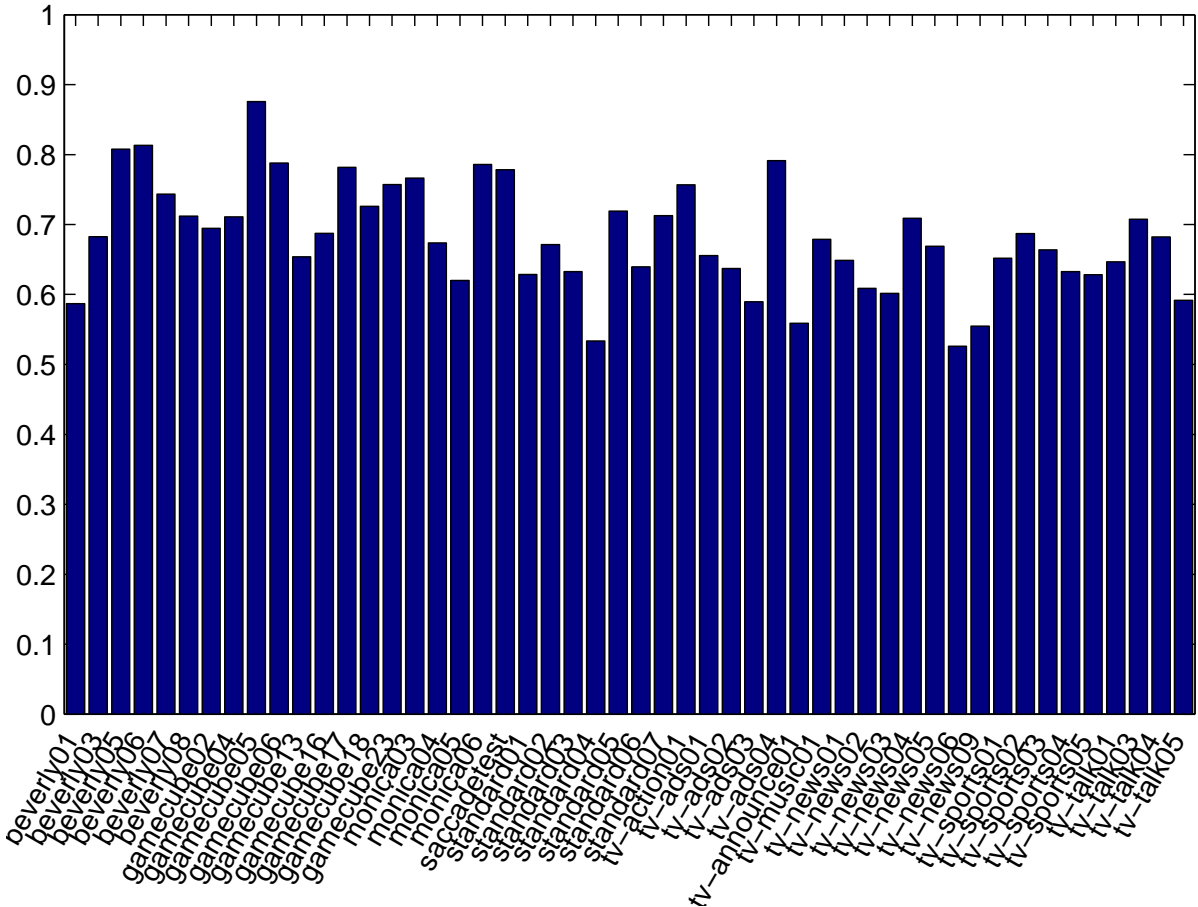}}
   \caption{The AUC of the proposed method for each video from CRCNS-ORIG dataset.}\label{fig:crcns}
\end{figure}

\begin{figure}[t]
  \centerline{\includegraphics[width=16cm, height=0.5\textheight]{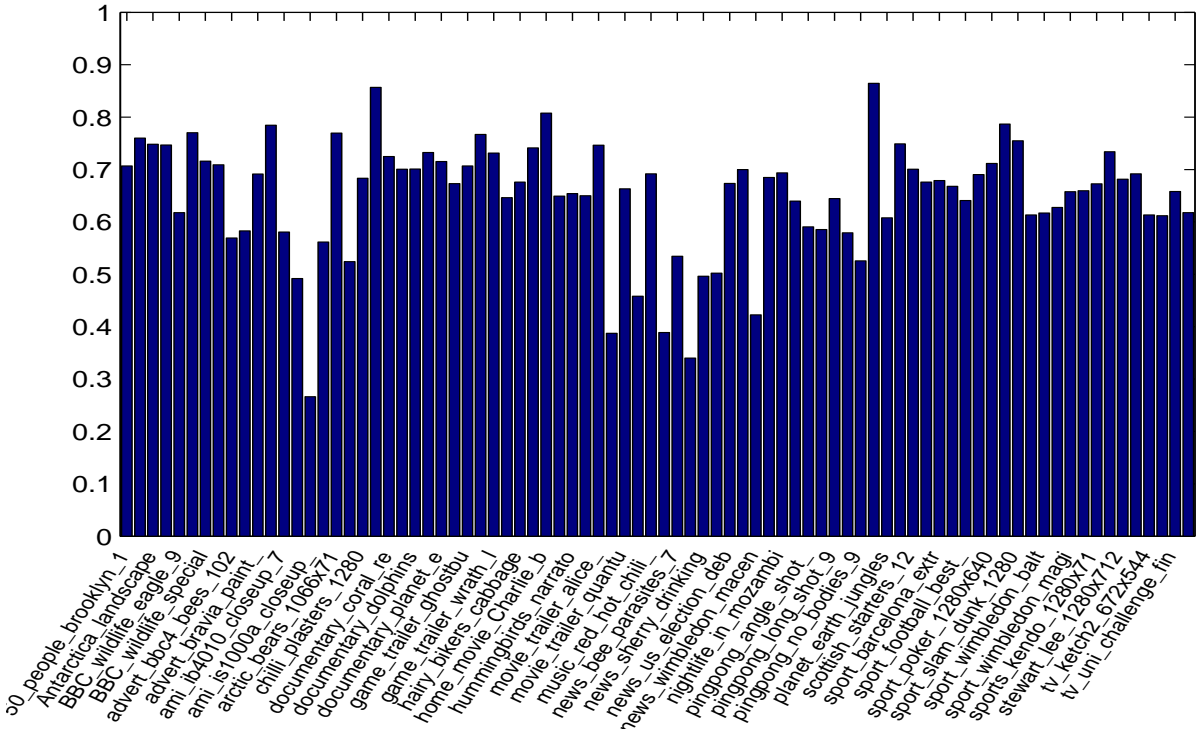}}
   \caption{The AUC of the proposed method for each video from DIEM dataset.}\label{fig:diem}
\end{figure}

We also include the AUC of the proposed method for each video from the CRCNS-ORIG (Fig. \ref{fig:crcns}) and DIEM dataset (Fig. \ref{fig:diem}).





\bibliographystyle{IEEE}
\bibliography{egbib_short}

\begin{thebibliography}{10}
\providecommand{\url}[1]{#1}
\csname url@samestyle\endcsname
\providecommand{\newblock}{\relax}
\providecommand{\bibinfo}[2]{#2}
\providecommand{\BIBentrySTDinterwordspacing}{\spaceskip=0pt\relax}
\providecommand{\BIBentryALTinterwordstretchfactor}{4}
\providecommand{\BIBentryALTinterwordspacing}{\spaceskip=\fontdimen2\font plus
\BIBentryALTinterwordstretchfactor\fontdimen3\font minus
  \fontdimen4\font\relax}
\providecommand{\BIBforeignlanguage}[2]{{%
\expandafter\ifx\csname l@#1\endcsname\relax
\typeout{** WARNING: IEEEtran.bst: No hyphenation pattern has been}%
\typeout{** loaded for the language `#1'. Using the pattern for}%
\typeout{** the default language instead.}%
\else
\language=\csname l@#1\endcsname
\fi
#2}}
\providecommand{\BIBdecl}{\relax}
\BIBdecl

\bibitem{itti1998model}
L.~Itti, C.~Koch, and E.~Niebur, ``A model of saliency-based visual attention
  for rapid scene analysis,'' \emph{PAMI}, vol.~20, no.~11, pp. 1254 --1259,
  nov 1998.

\bibitem{borji2012state}
A.~Borji and L.~Itti, ``State-of-the-art in visual attention modeling,''
  \emph{PAMI}, vol.~PP, no.~99, p.~1, 2012.

\bibitem{Alexe2012objectness}
B.~Alexe, T.~Deselaers, and V.~Ferrari, ``Measuring the objectness of image
  windows,'' \emph{PAMI}, vol.~34, no.~11, pp. 2189 --2202, nov. 2012.

\bibitem{sharma2012discriminative}
G.~Sharma, F.~Jurie, and C.~Schmid, ``Discriminative spatial saliency for image
  classification,'' in \emph{CVPR 2010}, june 2012, pp. 3506 --3513.

\bibitem{li2011saliency}
Q.~Li, Y.~Zhou, and J.~Yang, ``Saliency based image segmentation,'' in
  \emph{Multimedia Technology (ICMT), 2011 International Conference on}, july
  2011, pp. 5068 --5071.

\bibitem{hou2007saliency}
X.~Hou and L.~Zhang, ``Saliency detection: A spectral residual approach,'' in
  \emph{CVPR 2007}, june 2007, pp. 1 --8.

\bibitem{guo2008spatio}
C.~Guo, Q.~Ma, and L.~Zhang, ``Spatio-temporal saliency detection using phase
  spectrum of quaternion fourier transform,'' in \emph{CVPR 2008}, june 2008,
  pp. 1 --8.

\bibitem{hou2012image}
X.~Hou, J.~Harel, and C.~Koch, ``Image signature: Highlighting sparse salient
  regions,'' \emph{PAMI}, vol.~34, no.~1, pp. 194--201, jan. 2012.

\bibitem{ma2005generic}
Y.-F. Ma, X.-S. Hua, L.~Lu, and H.-J. Zhang, ``A generic framework of user
  attention model and its application in video summarization,''
  \emph{Multimedia, IEEE Transactions on}, vol.~7, no.~5, pp. 907 -- 919, oct.
  2005.

\bibitem{itti2004realistic}
L.~Itti, N.~Dhavale, and F.~Pighin, ``Realistic avatar eye and head animation
  using a neurobiological model of visual attention,'' in \emph{Optical Science
  and Technology, SPIE's 48th Annual Meeting}.\hskip 1em plus 0.5em minus
  0.4em\relax International Society for Optics and Photonics, 2004, pp. 64--78.

\bibitem{guo2010novel}
C.~Guo and L.~Zhang, ``A novel multiresolution spatiotemporal saliency
  detection model and its applications in image and video compression,''
  \emph{Image Processing, IEEE Transactions on}, vol.~19, no.~1, pp. 185 --198,
  jan. 2010.

\bibitem{gao2009discriminant}
D.~Gao, S.~Han, and N.~Vasconcelos, ``Discriminant saliency, the detection of
  suspicious coincidences, and applications to visual recognition,''
  \emph{PAMI}, vol.~31, no.~6, pp. 989 --1005, june 2009.

\bibitem{olveczky2003segregation}
B.~{\"O}lveczky, S.~Baccus, and M.~Meister, ``Segregation of object and
  background motion in the retina,'' \emph{Nature}, vol. 423, no. 6938, pp.
  401--408, 2003.

\bibitem{bian2009biological}
P.~Bian and L.~Zhang, ``Biological plausibility of spectral domain approach for
  spatiotemporal visual saliency,'' \emph{NIPS}, pp. 251--258, 2009.

\bibitem{simoncelli99modelingsurround}
E.~P. Simoncelli and O.~Schwartz, ``Modeling surround suppression in v1 neurons
  with a statistically-derived normalization model,'' 1999.

\bibitem{zhaoping2006pre}
L.~Zhaoping and P.~Dayan, ``Pre-attentive visual selection,'' \emph{Neural
  Networks}, vol.~19, no.~9, pp. 1437--1439, 2006.

\bibitem{huber2005visualizing}
D.~E. Huber and C.~G. Healey, ``Visualizing data with motion,'' in
  \emph{Visualization, 2005. VIS 05. IEEE}.\hskip 1em plus 0.5em minus
  0.4em\relax IEEE, 2005, pp. 527--534.

\bibitem{zhai2006visual}
\BIBentryALTinterwordspacing
Y.~Zhai and M.~Shah, ``Visual attention detection in video sequences using
  spatiotemporal cues,'' in \emph{Proceedings of the 14th ACM Multimedia}, ser.
  MULTIMEDIA '06.\hskip 1em plus 0.5em minus 0.4em\relax New York, NY, USA:
  ACM, 2006, pp. 815--824. [Online]. Available:
  \url{http://doi.acm.org/10.1145/1180639.1180824}
\BIBentrySTDinterwordspacing

\bibitem{zhang2009sunday}
L.~Zhang, M.~Tong, and G.~Cottrell, ``Sunday: Saliency using natural statistics
  for dynamic analysis of scenes,'' in \emph{Proceedings of the 31st Annual
  Cognitive Science Conference}.\hskip 1em plus 0.5em minus 0.4em\relax AAAI
  Press Cambridge, MA, 2009, pp. 2944--2949.

\bibitem{seo2011static}
\BIBentryALTinterwordspacing
H.~J. Seo and P.~Milanfar, ``Static and space-time visual saliency detection by
  self-resemblance,'' \emph{Journal of Vision}, vol.~9, no.~12, 2009. [Online].
  Available: \url{http://www.journalofvision.org/content/9/12/15.abstract}
\BIBentrySTDinterwordspacing

\bibitem{li2010visual}
Y.~Li, Y.~Zhou, J.~Yan, Z.~Niu, and J.~Yang, ``Visual saliency based on
  conditional entropy,'' \emph{ACCV 2009}, pp. 246--257, 2010.

\bibitem{mahadevan2010spatio}
V.~Mahadevan and N.~Vasconcelos, ``Spatiotemporal saliency in dynamic scenes,''
  \emph{PAMI}, vol.~32, no.~1, pp. 171--177, 2010.

\bibitem{ban2008dynamic}
S.~Ban, I.~Lee, and M.~Lee, ``Dynamic visual selective attention model,''
  \emph{Neurocomputing}, vol.~71, no.~4, pp. 853--856, 2008.

\bibitem{itti2009crcns}
\BIBentryALTinterwordspacing
R.~Itti, Laurent;~Carmi, ``Eye-tracking data from human volunteers watching
  complex video stimuli,'' Online, 2009. [Online]. Available: \url{CRCNS.org}
\BIBentrySTDinterwordspacing

\bibitem{mital2011clustering}
P.~K. Mital, T.~J. Smith, R.~L. Hill, and J.~M. Henderson, ``Clustering of gaze
  during dynamic scene viewing is predicted by motion,'' \emph{Cognitive
  Computation}, vol.~3, no.~1, pp. 5--24, 2011.

\bibitem{garcia2009decorrelation}
A.~Garcia-Diaz, X.~R. Fdez-Vidal, X.~M. Pardo, and R.~Dosil, ``Decorrelation
  and distinctiveness provide with human-like saliency,'' in \emph{Advanced
  Concepts for Intelligent Vision Systems}.\hskip 1em plus 0.5em minus
  0.4em\relax Springer, 2009, pp. 343--354.

\bibitem{hou2008dynamic}
X.~Hou and L.~Zhang, ``Dynamic visual attention: Searching for coding length
  increments,'' \emph{NIPS}, vol.~21, pp. 681--688, 2008.

\bibitem{marat2009modelling}
S.~Marat, T.~Ho~Phuoc, L.~Granjon, N.~Guyader, D.~Pellerin, and
  A.~Gu{\'e}rin-Dugu{\'e}, ``Modelling spatio-temporal saliency to predict gaze
  direction for short videos,'' \emph{IJCV}, vol.~82, no.~3, pp. 231--243,
  2009.

\bibitem{judd2009learning}
T.~Judd, K.~Ehinger, F.~Durand, and A.~Torralba, ``Learning to predict where
  humans look,'' in \emph{ICCV 2009}, 29 2009-oct. 2 2009, pp. 2106 --2113.

\bibitem{bruce2005saliency}
N.~Bruce and J.~Tsotsos, ``Saliency based on information maximization,'' in
  \emph{Advances in neural information processing systems}, 2005, pp. 155--162.

\bibitem{torralba2003modeling}
A.~Torralba, ``Modeling global scene factors in attention,'' \emph{JOSA A},
  vol.~20, no.~7, pp. 1407--1418, 2003.

\bibitem{harel2006graph}
J.~Harel, C.~Koch, and P.~Perona, ``Graph-based visual saliency,'' in
  \emph{Advances in neural information processing systems}, 2006, pp. 545--552.

\bibitem{mancas2007computational}
M.~Mancas, ``Computational attention: Modelisation and application to audio and
  image processing,'' Ph.D. dissertation, PhD. Thesis, University of Mons,
  2007.

\bibitem{itti2006bayesian}
L.~Itti and P.~Baldi, ``Bayesian surprise attracts human attention,''
  \emph{NIPS}, vol.~18, p. 547, 2006.

\bibitem{borji2012quantitative}
A.~Borji, D.~N. Sihite, and L.~Itti, ``Quantitative analysis of human-model
  agreement in visual saliency modeling: a comparative study,'' 2012.

\bibitem{mahadevan2010anomaly}
V.~Mahadevan, W.~Li, V.~Bhalodia, and N.~Vasconcelos, ``Anomaly detection in
  crowded scenes,'' in \emph{CVPR 2010}, june 2010, pp. 1975 --1981.

\bibitem{mehran2009abnormal}
R.~Mehran, A.~Oyama, and M.~Shah, ``Abnormal crowd behavior detection using
  social force model,'' in \emph{CVPR 2009}, june 2009, pp. 935 --942.

\bibitem{cong2011sparse}
Y.~Cong, J.~Yuan, and J.~Liu, ``Sparse reconstruction cost for abnormal event
  detection,'' in \emph{CVPR 2011}, june 2011, pp. 3449 --3456.

\bibitem{kim2009observe}
J.~Kim and K.~Grauman, ``Observe locally, infer globally: A space-time mrf for
  detecting abnormal activities with incremental updates,'' in \emph{CVPR
  2009}, june 2009, pp. 2921 --2928.

\bibitem{wu2010chaotic}
S.~Wu, B.~Moore, and M.~Shah, ``Chaotic invariants of lagrangian particle
  trajectories for anomaly detection in crowded scenes,'' in \emph{CVPR 2010},
  june 2010, pp. 2054 --2060.

\bibitem{raghavendra2011optimizing}
R.~Raghavendra, A.~Del~Bue, M.~Cristani, and V.~Murino, ``Optimizing
  interaction force for global anomaly detection in crowded scenes,'' in
  \emph{Computer Vision Workshops (ICCV Workshops), 2011 IEEE International
  Conference on}, nov. 2011, pp. 136 --143.

\bibitem{adam2008robust}
A.~Adam, E.~Rivlin, I.~Shimshoni, and D.~Reinitz, ``Robust real-time unusual
  event detection using multiple fixed-location monitors,'' \emph{PAMI},
  vol.~30, no.~3, pp. 555 --560, march 2008.

\bibitem{reddy2011improved}
V.~Reddy, C.~Sanderson, and B.~Lovell, ``Improved anomaly detection in crowded
  scenes via cell-based analysis of foreground speed, size and texture,'' in
  \emph{Computer Vision and Pattern Recognition Workshops (CVPRW), 2011 IEEE
  Computer Society Conference on}, june 2011, pp. 55 --61.

\bibitem{laptev2005space}
I.~Laptev, ``On space-time interest points,'' \emph{IJCV}, vol.~64, no.~2, pp.
  107--123, 2005.

\bibitem{lena2007actions}
L.~Gorelick, M.~Blank, E.~Shechtman, M.~Irani, and R.~Basri, ``Actions as
  space-time shapes,'' \emph{PAMI}, vol.~29, no.~12, pp. 2247--2253, December
  2007.

\bibitem{schuldt2004recognizing}
C.~Schuldt, I.~Laptev, and B.~Caputo, ``Recognizing human actions: a local svm
  approach,'' in \emph{ICPR 2004}, vol.~3, aug. 2004, pp. 32 -- 36 Vol.3.

\bibitem{rodriguez2008action}
M.~Rodriguez, J.~Ahmed, and M.~Shah, ``Action mach a spatio-temporal maximum
  average correlation height filter for action recognition,'' in \emph{CVPR
  2008}, june 2008, pp. 1--8.

\bibitem{dollar2005behavior}
P.~Dollar, V.~Rabaud, G.~Cottrell, and S.~Belongie, ``Behavior recognition via
  sparse spatio-temporal features,'' in \emph{Visual Surveillance and
  Performance Evaluation of Tracking and Surveillance, 2005. 2nd Joint IEEE
  International Workshop on}, oct. 2005, pp. 65 -- 72.

\bibitem{willems2008efficient}
G.~Willems, T.~Tuytelaars, and L.~Van~Gool, ``An efficient dense and
  scale-invariant spatio-temporal interest point detector,'' \emph{ECCV 2008},
  pp. 650--663, 2008.

\bibitem{kla2010learning}
\BIBentryALTinterwordspacing
A.~Kl{\"a}ser, ``Learning human actions in video,'' Ph.D. dissertation,
  Universit\'e de Grenoble, jul 2010. [Online]. Available:
  \url{http://lear.inrialpes.fr/pubs/2010/Kla10}
\BIBentrySTDinterwordspacing

\end{thebibliography}
%

%








\end{document}